\documentclass{article}
\usepackage[preprint]{spconf}
\usepackage{spconf,amsmath,graphicx}
\usepackage{tabulary}
\usepackage{xcolor}
\usepackage{xspace}
\usepackage{booktabs}
\usepackage{makecell}
\usepackage{multirow}
\usepackage{textcomp}
\usepackage{inputenc}
\usepackage{xurl}
\usepackage[font=normalsize,skip=12pt]{caption,subcaption}

\newlength\figwidth
\newlength\imagewidth
\newlength\subcap

\makeatletter
\copyrightnotice{
  {\footnotesize
  \begin{minipage}{\textwidth}
  \centering
  Copyright~\copyright~2023 IEEE.  Personal use of this material is permitted.  Permission from IEEE must be obtained for all other uses, in any current or future media, including reprinting/republishing this material for advertising or promotional purposes, creating new collective works, for resale or redistribution to servers or lists, or reuse of any copyrighted component of this work in other works.
  \end{minipage}
  }
}
\DeclareRobustCommand\onedot{\futurelet\@let@token\@onedot}
\def\@onedot{\ifx\@let@token.\else.\null\fi\xspace}

\def\eg{\emph{e.g}\onedot}

\def\etal{et al\onedot}
\makeatother

\title{Advancing the Rate-Distortion-Computation Frontier\\for Neural Image Compression}
\name{David Minnen \& Nick Johnston}
\address{Google Research, Mountain View, CA 94043, USA}
\begin{document}
%
\maketitle
\begin{abstract}
The rate-distortion performance of neural image compression models has exceeded the state-of-the-art for non-learned codecs, but neural codecs are still far from widespread deployment and adoption. The largest obstacle is having efficient models that are feasible on a wide variety of consumer hardware. Comparative research and evaluation is difficult due to the lack of standard benchmarking platforms and due to variations in hardware architectures and test environments. Through our rate-distortion-computation (RDC) study we demonstrate that neither floating-point operations (FLOPs) nor runtime are sufficient on their own to accurately rank neural compression methods.
We also explore the RDC frontier, which leads to a family of model architectures with the best empirical trade-off between computational requirements and RD performance. Finally, we identify a novel neural compression architecture that yields state-of-the-art RD performance with rate savings of 23.1\% over BPG (7.0\% over VTM and 3.0\% over ELIC) without requiring significantly more FLOPs than other learning-based codecs.
\end{abstract}
%

\section{Introduction}
\label{sec:intro}
The best neural image compression models are able to outperform leading non-learned image codecs in  rate-distortion (RD)
performance. However, neural image compression typically requires significantly more decode
time and powerful, specialized hardware to achieve feasible runtimes for real-world applications.
So far, no standard for providing runtime characteristics in neural image compression research 
has been adopted. While some studies report FLOPs, others report runtime on a
particular device. We will demonstrate that neither of these alone is sufficient to meaningfully compare with other research works.

In this paper, we first show the results of a large architecture and parameter sweep over common
neural image compression architectures. Next, we show how decoder runtime and decoder FLOPs (shortened to simply \textit{runtime} and \textit{FLOPs} below) are correlated across
different model architectures and hardware platforms. Then, using the architecture study, we highlight a family of models that empirically have the best trade-off between FLOPs and RD performance. This family includes a new state-of-the-art model compared to BPG (HEVC Still Picture format), VTM (the VVC test model), and previous neural codecs. Finally, we make recommendations on how researchers should report the
runtime characteristics of their models to more easily understand the trade-offs between compute, runtime, and RD performance.

\begin{figure}[t]
  \centering
  \includegraphics[width=\linewidth]{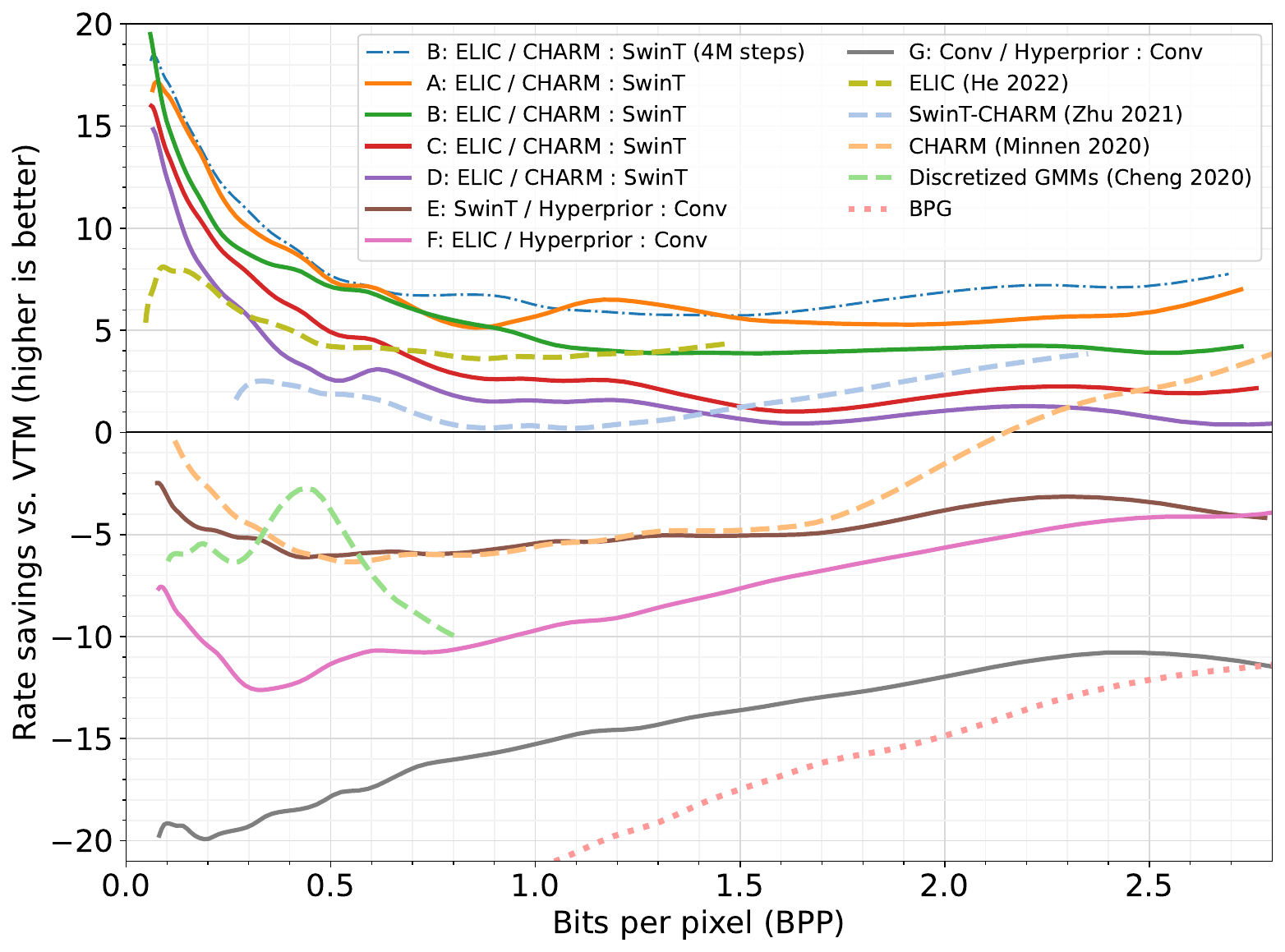}
  \caption{Our model (``B'') trained for 4M steps achieves state-of-the-art RD performance for MSE-optimized compression (see Fig.~\ref{fig:scatter-loss-vs-flops} for a FLOPs comparison). This graph shows rate savings relative to VTM v13.0 (higher is better) averaged over the Kodak dataset~\cite{kodak}.}
  \label{fig:rate-savings}
\end{figure}

\begin{figure*}[tb]
  \centering
  \includegraphics[width=\linewidth]{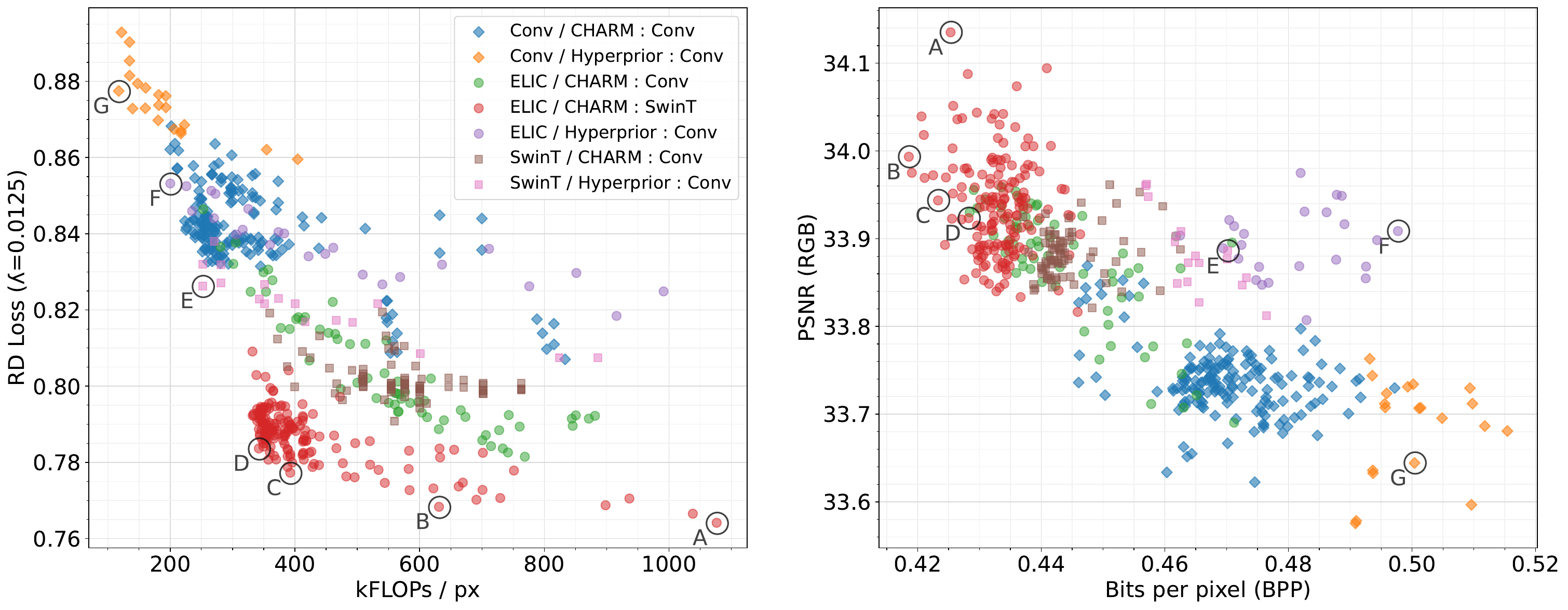}
  \caption{The left plot shows model quality (RD loss) vs. complexity (kFLOPs per pixel) for 571 architecture variations. The best models compress well (smaller RD loss) and require fewer FLOPs. The plot on the right shows the same data but visualizes reconstruction quality (PSNR) vs. bit rate (BPP). Each color represents an architecture family labeled according to the modeling choice for the three main components: $\langle$ analysis \& synthesis transforms$\rangle$ ~/ $\langle$ entropy model$\rangle$ ~: $\langle$ hyperprior transforms$\rangle$.
  }
  \label{fig:scatter-loss-vs-flops}
\end{figure*}

\section{Prior Work}
\label{sec:prior}
Our architecture study builds on several previous model innovations: a hyperprior for entropy modeling~\cite{balle2018iclr}, channel-wise autoregressive models (CHARM)~\cite{minnen2020icip}, ELIC's use of convolutional layers, residual blocks, and attention blocks for transforms~\cite{he2022elic}, and the Swin Transformer (SwinT) applied to compression models~\cite{zhu2022, zou2022devil, liu2021swin}.
Johnston~\etal produced a survey across different rates and loss functions, but focused on a single architecture~\cite{johnston2019cenic}. Yang~\etal ~\cite{yang2020improving} and Wu~\etal~\cite{wu2021latent} showed increased RD performance using instance-adaptive methods that significantly increase image encoding time but do not slow down decoding. Le~\etal demonstrates the importance of research and engineering co-design to optimize the runtime of end-to-end neural compression systems~\cite{le2022mobilecodec}. And Deghani~\etal demonstrate discrepancies between FLOPs, runtime, and parameter counts in computer vision models and highlight the importance of accurate reporting of multiple metrics~\cite{deghani2010}.

Previous research on neural image compression includes work that boosts RD performance without considering computational implications,~\eg, Cheng~\etal used Gaussian mixture models for better entropy models~\cite{cheng2020learned}, and several papers explored spatial autoregressive (AR) context models which can be very slow in practice since they do not effectively utilize massively parallel hardware~\cite{klopp2018bmvc, lee2019cae, minnen2018joint}. Other research puts more emphasis on efficiency. The channel-wise autoregressive model (CHARM) was developed to improve parallelization and reduce decode time compared to spatial AR context models~\cite{minnen2020icip}. He~\etal introduced ELIC, which improved CHARM by using uneven channel-wise groups and a spatial checkerboard decomposition~\cite{he2022elic}. And a different line of research explored vector quantization (VQ) for the latent representation~\cite{lu2019dvq,zhu2022multivariate} since it typically leads to faster entropy models than scalar quantization.

\section{Architecture Study}
\label{sec:survey}
We use L2 loss for distortion and a fixed RD trade-off of $\lambda = 0.0125$ which leads to an average of 0.46 bits per pixel on the Kodak image set used for evaluation~\cite{kodak} (see Fig.~\ref{fig:scatter-loss-vs-flops}).
We trained different combinations of analysis and synthesis transforms (stacked convolutions~\cite{balle2018iclr}, ELIC-style~\cite{he2022elic}, and SwinT~\cite{zhu2022,zou2022devil,liu2021swin}), entropy models (hyperprior~\cite{balle2018iclr} and CHARM~\cite{minnen2020icip}), and hyperprior transforms (stacked convolutions and SwinT) along with parameter sweeps across channel depth and number of layers. 
All models were trained using TensorFlow 
for 2M steps with a batch size of 32 (or 16 when memory constraints required it).

Fig.~\ref{fig:scatter-loss-vs-flops} (\textit{left}) shows the results of our architecture sweeps and highlights seven models along the RDC frontier (labeled A--G, see Fig.~\ref{fig:rate-savings} and Table~\ref{table:runtime} for details). Although we train with a single $\lambda$ to target a fixed RD trade-off, there is some variation in the final bit rates and reconstruction quality levels as shown in Fig.~\ref{fig:scatter-loss-vs-flops} (\textit{right}). Several interesting trends emerge: models with stacked convolutions (labeled as \textit{Conv}) generally lead to the lowest compute but worst RD loss. This is not surprising since there are only four deconvolution layers in the synthesis transform for these models. Moving from a basic hyperprior to CHARM for the entropy model increases both the complexity and quality of the model (\eg, compare orange vs. blue diamonds, purple vs. green dots, and pink vs. brown squares). Updating the transforms from stacked convolutions to ELIC makes the models more complex but also provides a significant gain in compression quality (compare blue diamonds to green dots). Comparing ELIC to SwinT for the transforms (green dots vs. brown squares) is more complex since relative RD performance flips for low vs. high FLOP counts. Finally, we see that using SwinT in the hyperprior leads to a more powerful entropy model with better RD performance across a wide complexity range (red dots).

\begin{figure}[tb]
  \centering
  \includegraphics[width=\linewidth]{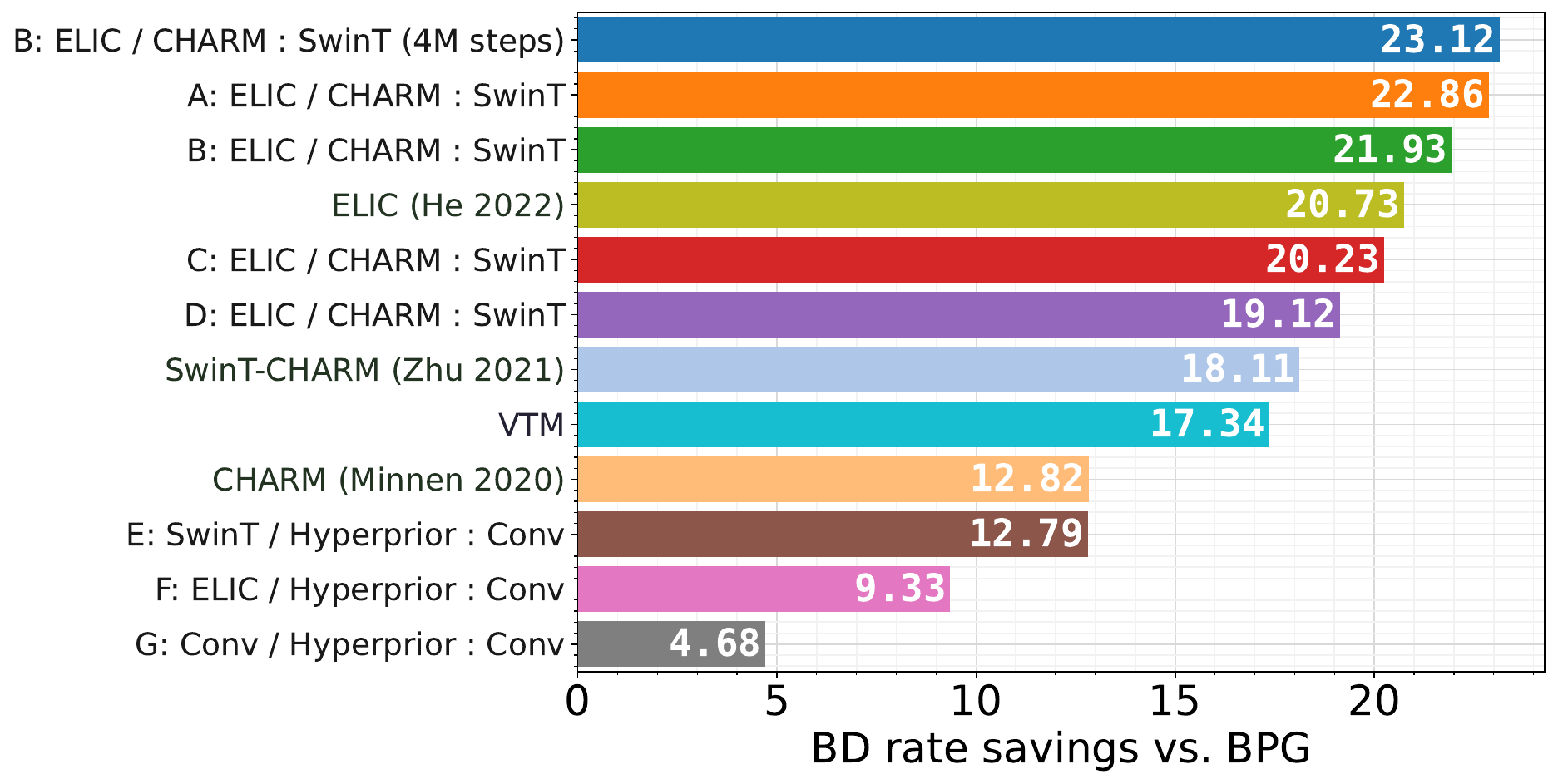}
  \caption{Bjøntegaard Delta (BD) chart~\cite{VCEG-M33} showing average rate savings compared to BPG~\cite{bpg}. All BD rates were calculated over the largest quality range covered by all models (31.6 dB -- 40.8 dB).
  For a more fine-grain comparison, Fig.~\ref{fig:rate-savings} shows how rate savings varies across bit rates.}
  \label{fig:bd-chart}
\end{figure}

\begin{figure*}[tb]
  \centering
  \includegraphics[width=0.32\linewidth]{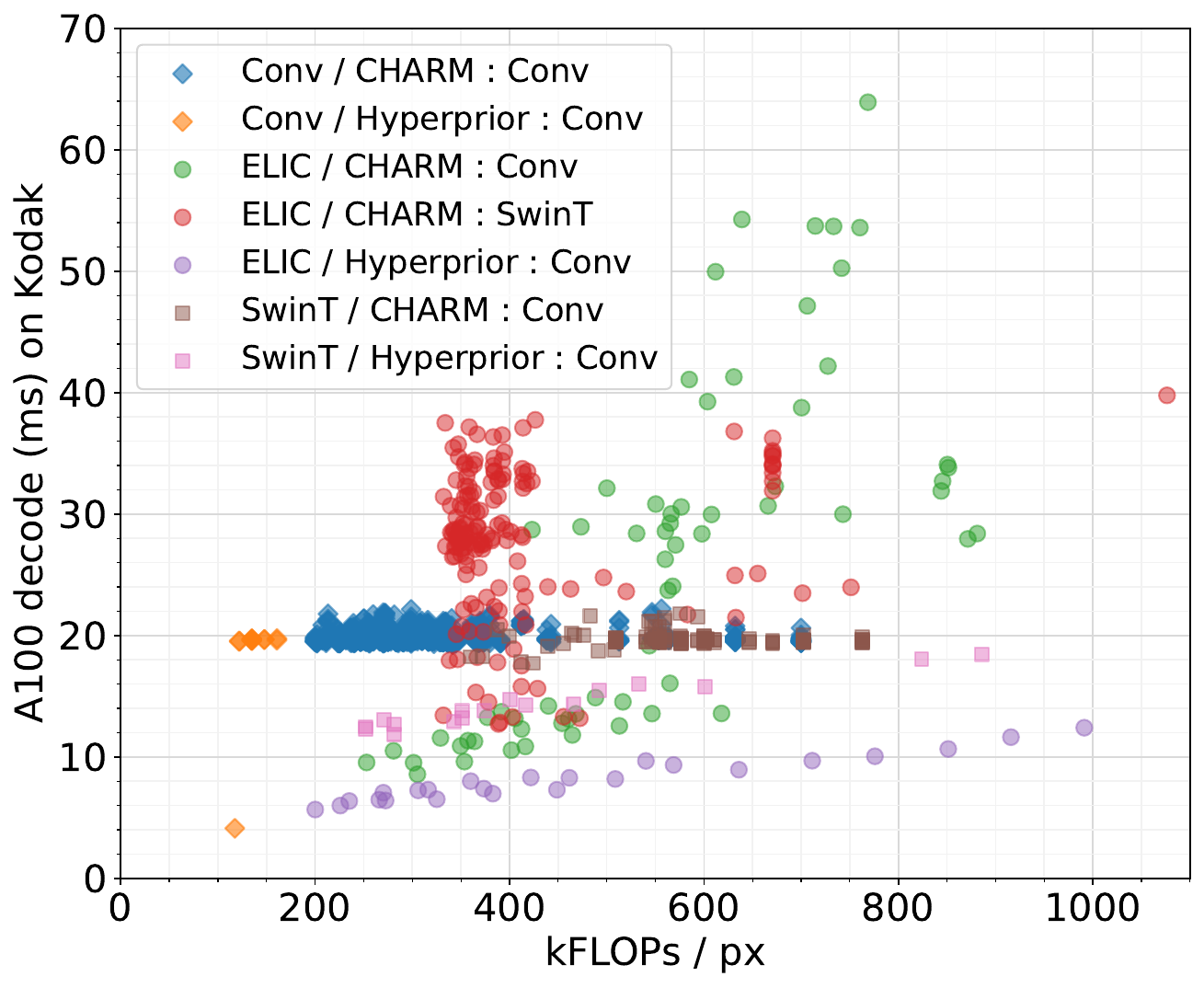}
  \hfill
  \includegraphics[width=0.34\linewidth]{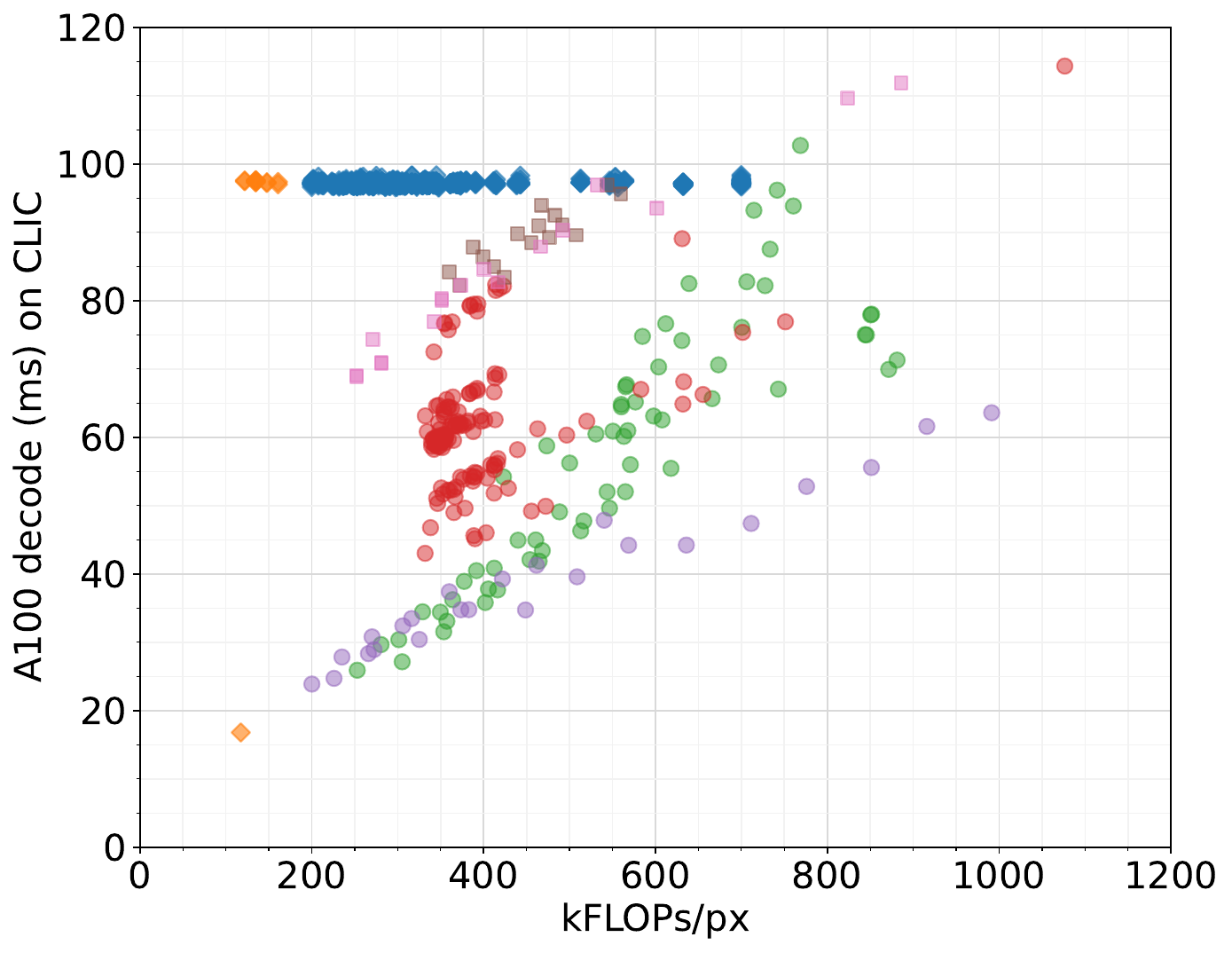}
  \hfill
  \includegraphics[width=0.33\linewidth]{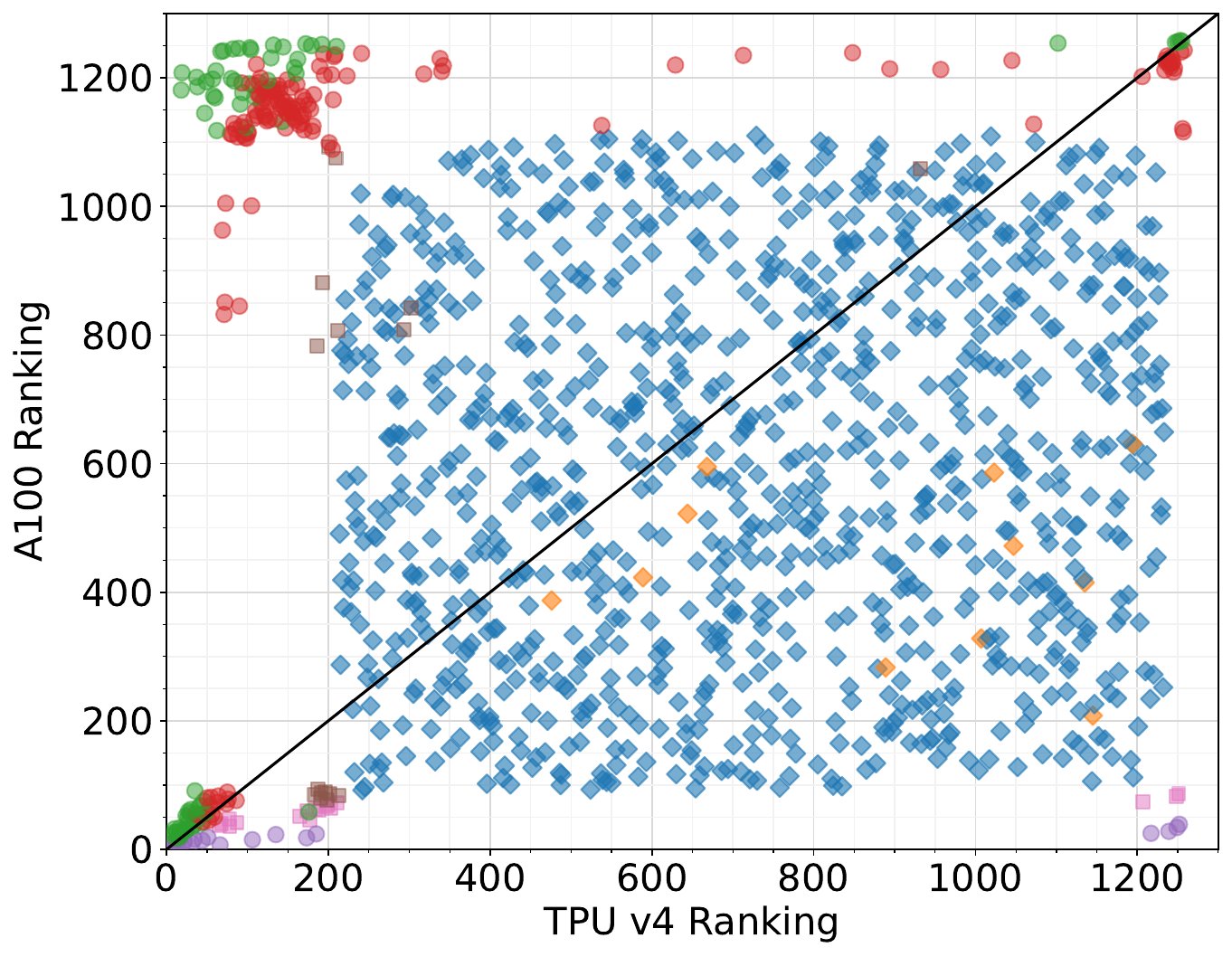}
  
  \vspace{-2pt}
  \caption{Decode time vs. FLOPs (averaged over Kodak (\textit{left}) and over sixteen CLIC images (\textit{center})) shows a mostly linear relationship for models using a simple hyperprior~\cite{balle2018iclr} (purple dots, pink squares, and orange diamonds). This trend breaks down for CHARM-based entropy models~\cite{minnen2020icip} where we see virtually no speed-up despite a 50\% reduction in FLOPS (blue diamonds) and see wide runtime variance despite consistent FLOPs (red circles). We rank our methods fastest to slowest and compare rankings between an A100 GPU and a TPU v4 (\textit{right}). Models below the diagonal line run relatively better on an A100, and models above the line run relatively better on a TPU v4. This makes ranking runtimes difficult, as it is very unlikely one can reproduce the relative ordering of methods on the particular hardware available to that researcher. This plot also reinforces the importance of hardware-software co-design if one wants to optimize directly for a particular device's runtime.}
  \label{fig:scatter-runtime-vs-flops}
  \vspace{7pt}  
\end{figure*}

Many of the interesting points on the frontier correspond to models that use ELIC transforms, a CHARM entropy model, and SwinT hyper-transforms. This architecture provides a strong baseline for further research into high-quality neural image compression but is at least 4x slower than smaller models that use simpler stacked convolutions and a basic hyperprior. To better evaluate the frontier models, relative rate savings are shown in Fig.~\ref{fig:rate-savings} \&~\ref{fig:bd-chart}, and runtime information is  provided in Fig.~\ref{fig:scatter-runtime-vs-flops} and Table~\ref{table:runtime}.


\section{Relationship of FLOPs to Runtime}
\label{sec:flops_to_runtime}
The FLOPs used by a neural model is not always proportional to runtime~\cite{deghani2010}. Several factors come into play here including the computational architecture, hardware and software used, and even the environment. The computational architecture, or how the FLOPs are arranged, is one factor that can affect runtime parallelism. For example, we can have one largely parallelized matrix multiplication that may complete in one time unit on a GPU, or we could have five very small matrix multiplications that cannot be fused together and therefore need to be computed serially. In this simple example, the one large operation might have more FLOPs yet still run more quickly than the five smaller operations.

Whether it's a TPU, GPU, FPGA or CPU, the design of each hardware platform can be optimized for certain operations or network architectures. Hardware availability, end-of-life, machine learning libraries, CUDA drivers, and even the operating system can affect the availability and performance of benchmarking. To provide more detailed information for our architecture study, we explored runtimes on a Nvidia A100 GPU~\cite{a100} and a Google TPU v4~\cite{tpuv4} for models on the empirical RDC frontier.


In Fig.~\ref{fig:scatter-runtime-vs-flops} we see our models plotted with kFLOPs per pixel on the x-axis and decode time on an A100 GPU on the y-axis averaged over Kodak (\textit{left}) and a 16 image CLIC subset (\textit{center}). We provide timing data without entropy coding since optimizing the entropy coder is not the focus of this paper. Our unoptimized entropy coder leads to a relatively constant overhead across models and thus masks runtime differences due to changes in the neural architecture.
If FLOPs is a good indicator of runtime, there should be a strong linear correlation in these scatterplots. While we do see linear correlation for some architectures, others show a very different relationship, \eg, the {\small\texttt{Conv/CHARM:Conv}} (blue diamonds) and {\small\texttt{Conv/Hyperprior:Conv}} models (orange diamonds) have a very strong grouping around 20 ms. This is likely due to the A100 being under utilized. Thus, increasing the compute (by more than 6x in our study) added FLOPs in a way that was parallelizable with existing operations and thus did not have a linear effect on runtime.
In general, the models with a CHARM entropy model show that you can have small changes in the model's FLOPs but a large multiplier on runtime. The resolution of the images across the two datasets also did not change these trends. Thus, while FLOP counts are platform agnostic, they are not sufficient for comparing relative runtime across model architectures.

Finally, we wanted to compare the relative performance between the A100 and TPU v4 hardware platforms. In Fig.~\ref{fig:scatter-runtime-vs-flops} (\textit{right}), we rank the runtime of all of our models on an A100 GPU and on a TPU v4. We see that across accelerator platforms it is not guaranteed to maintain the same set of ranking for different models. None of the model architecture families we explored were immune to this device-dependent variation.

While our study shows that comparisons across architectures and across devices may lead to false equivalences in neural compression research, it is important to keep the goal in mind. Runtime on a particular device can be useful for determining which models are ``faster'', though the claim of speed may not hold outside of that device and environment. For researchers more interested in complexity, FLOPs is an indicator within a particular architecture
but can mask actual complexity when comparing across different architectures.

In the past, CLIC~\cite{clic2022} ran many methods in the same environment, but that environment still changed each year. A global, year-round leaderboard, like ImageNet~\cite{imagenet} and MS-COCO~\cite{mscoco}, would allow compression researchers to consistently compare progress over time. In the absence of a persistent leaderboard for measuring RD and runtime data, both FLOPs and runtime on device should be reported along with a more detailed discussion on how those FLOPs are distributed when they are compared across different architectures.

\newcommand\mc[1]{\multicolumn{2}{c}{#1}}
\newcommand\mr[1]{\multirow{2}{*}{#1}}
\begin{table*}[tb]
  \centering
  \footnotesize
  \setlength\extrarowheight{2pt}
  \setlength\tabcolsep{6pt}
\begin{tabulary}{\textwidth}{@{}c|ccc|c|c|cc|cc|c@{}}
\toprule
   \mr{\thead{Model}}
 & \mr{\thead{Transforms\\ (ana/syn)}}
 & \mr{\thead{Entropy\\Model}}
 & \mr{\thead{Hyperprior\\Transforms}}
 & \mr{\thead{Parameters $\downarrow$\\ \scriptsize{(millions)}}}
 & \mr{\thead{kFLOPS/px $\downarrow$ \\ \scriptsize{(decode-only)}}}
 & \multicolumn{2}{c|}{\thead{Decode MP/s (Kodak) $\uparrow$}}
 & \multicolumn{2}{c|}{\thead{Decode MP/s (CLIC) $\uparrow$}}
 & \mr{\thead{Rate Savings $\uparrow$ \\ vs. BPG}}
 \\
  &  &  &  &  &  & V100 & A100 & V100 & A100 &
 \\ \hline
A & ELIC & CHARM & SwinT & 38.0 & 1076.4 &  7.4 &   9.9 & 10.1 &  27.5 & 22.86\\ 
B & ELIC & CHARM & SwinT & 31.6 &  631.1 &  8.8 &  10.7 & 14.8 &  35.3 & 21.93 \\ 
C & ELIC & CHARM & SwinT & 32.2 &  392.5 & 11.0 &  10.8 & 18.2 &  40.1 & 20.23 \\ 
D & ELIC & CHARM & SwinT & 22.4 &  342.1 & 11.7 &  11.1 & 19.6 &  43.4 & 19.12 \\ 
E & SwinT & Hyperprior & Conv & 17.3 &  252.1 & 20.5 &  31.5 & 23.9 &  45.5 & 12.79 \\ 
F & ELIC & Hyperprior & Conv & 19.8 &  200.1 & 41.0 &  69.0 & 51.7 & 131.6 & 9.33 \\ 
G & Conv & Hyperprior & Conv & 12.8 &  117.4 & 43.7 &  95.9 & 44.3 & 187.2 & 4.68 \\ 
\bottomrule
  \end{tabulary}
  \vspace{-2pt}
  \caption{Each row represents a model on the RDC frontier (see Fig.~\ref{fig:scatter-loss-vs-flops}) listed in descending order of complexity. Columns 2--4 describe the model architecture, while the remaining columns list the number of parameters, FLOPs per pixel, decode speed in megapixels per second, and rate savings over BPG. Each Kodak image has $512 \times 768 = 393{,}216$ pixels, and the runtime numbers for the CLIC 2021 Professional dataset are averaged over 16 images each with $2048 \times 1365 = 2{,}795{,}520$ pixels.}
  \label{table:runtime}
\end{table*}

\section{Improved RDC Models}
\label{sec:rdc_models}

The architecture that gave the best RD results with moderate complexity (``B'' in Table~\ref{table:runtime}) used ELIC-style analysis and synthesis transforms~\cite{he2022elic} with a CHARM-based entropy model~\cite{minnen2020icip}. We found that two changes to the published ELIC transforms improved the RDC trade-off in our model: first, we reduced the number of residual blocks from three to two, and second we changed the channel depths in the analysis transform from [192, 192, 192] to [128, 256, 256] (with an equivalent change for the synthesis transform). 

We also modified CHARM in three key ways. First, we found that moving from stacked convolutions to SwinT layers in the hyperprior improved RD performance. Second, we changed how the latent residual prediction (LRP) values were merged with decoded latents by switching from addition to concatenation followed by a $1 \times 1$ convolutional layer. And third, following Zhu et al. (see Fig. 12 in Appendix A.1 of~\cite{zhu2022}), we transformed the output of the hyper-synthesis transform into $N$ smaller tensors using convolutional layers, where $N$ is the number of channel-wise AR steps (following~\cite{minnen2020icip,zhu2022}, $N = 10$ for all of our models). Although this adds $N$ extra layers to the decoding network, these layers can run in parallel, and the total computation is reduced due to using smaller tensors when predicting entropy parameters and LRP values.


\section{Conclusion}
\label{sec:conclusion}

Ranking the performance of neural image compression models is difficult due to the inversions demonstrated by our runtime vs. FLOPs plots and by the variability in runtime across platforms. We recommend that both FLOPs and runtime be reported, and ideally open-source implementations are also provided to facilitate comparisons by other researchers. We will have our best model open-sourced at \url{https://github.com/tensorflow/compression/tree/master/models/rdc}.

The result of our study is a model with state-of-the-art RD performance that falls on the RDC frontier of our architecture exploration and which outperforms BPG, VTM, ELIC and SwinT-CHARM baselines. We are optimistic that the compression community can create more efficient and effective models, evaluation methods, and hardware to make neural image compression feasible for widespread deployment.



\clearpage
\bibliographystyle{IEEEbib}
\bibliography{main.bib}

\end{document}